\documentclass[11pt,a4paper]{article}
\usepackage[hyperref]{acl2020}
\usepackage{times}
\usepackage{latexsym}

\usepackage{microtype}
\usepackage{times}
\usepackage{latexsym}
\usepackage{verbatim}
\usepackage{url}
\usepackage{multirow}
\usepackage{balance}
\usepackage{subfigure}
\usepackage{latexsym}
\usepackage{amsmath}
\usepackage{amssymb}
\usepackage{caption}
\usepackage{graphicx}
\usepackage{url}
\usepackage{multicol}
\usepackage{multirow}
\usepackage{booktabs}
\usepackage{color}
\usepackage{subfigure}
\usepackage{array}
\usepackage{balance}
\usepackage{appendix}
\aclfinalcopy 
\def\aclpaperid{707} 

\definecolor{midnightgreen}{rgb}{0.0, 0.29, 0.33}
\definecolor{orange}{RGB}{255,127,0}

\title{Grounded Conversation Generation as Guided Traverses in Commonsense Knowledge Graphs}

\author{Houyu Zhang$^{1}$ \thanks{ \ \ Indicates equal contribution.} \thanks{ \ \ Part of work is conducted at Tsinghua University.}\qquad Zhenghao Liu$^{2*}$ \qquad Chenyan Xiong$^3$ \qquad Zhiyuan Liu$^2$ \\ 
$^1$Department of Computer Science, Brown University, Providence, USA\\
$^2$Department of Computer Science and Technology, Tsinghua University, Beijing, China\\
Institute for Artificial Intelligence, Tsinghua University, Beijing, China\\
State Key Lab on Intelligent Technology and Systems, Tsinghua University, Beijing, China\\
$^3$Microsoft Research AI,  Redmond, USA\\}

\date{}

\begin{document}

\maketitle

\begin{abstract}
Human conversations naturally evolve around related concepts and scatter to multi-hop concepts. This paper presents a new conversation generation model, ConceptFlow, which leverages commonsense knowledge graphs to explicitly model conversation flows. By grounding conversations to the concept space, ConceptFlow represents the potential conversation flow as traverses in the concept space along commonsense relations. The traverse is guided by graph attentions in the concept graph, moving towards more meaningful directions in the concept space, in order to generate more semantic and informative responses. Experiments on Reddit conversations demonstrate ConceptFlow's effectiveness over previous knowledge-aware conversation models and GPT-2 based models while using 70\% fewer parameters, confirming the advantage of explicit modeling conversation structures. All source codes of this work are available at \url{https://github.com/thunlp/ConceptFlow}.
\end{abstract}
\section{Introduction}
The rapid advancements of language modeling and natural language generation (NLG) techniques have enabled fully data-driven conversation models, which directly generate natural language responses for conversations~\cite{shang2015neural,vinyals2015neural,li2016deep}.
However, it is a common problem that the generation models may degenerate dull and repetitive contents~\cite{holtzman2019curious, welleck2019neural}, which, in conversation assistants, leads to off-topic and useless responses.~\cite{tang2019target, zhang2018generating, gao2019jointly}.

\begin{figure}[t]
 \centering
 \includegraphics[width=1.0\linewidth]{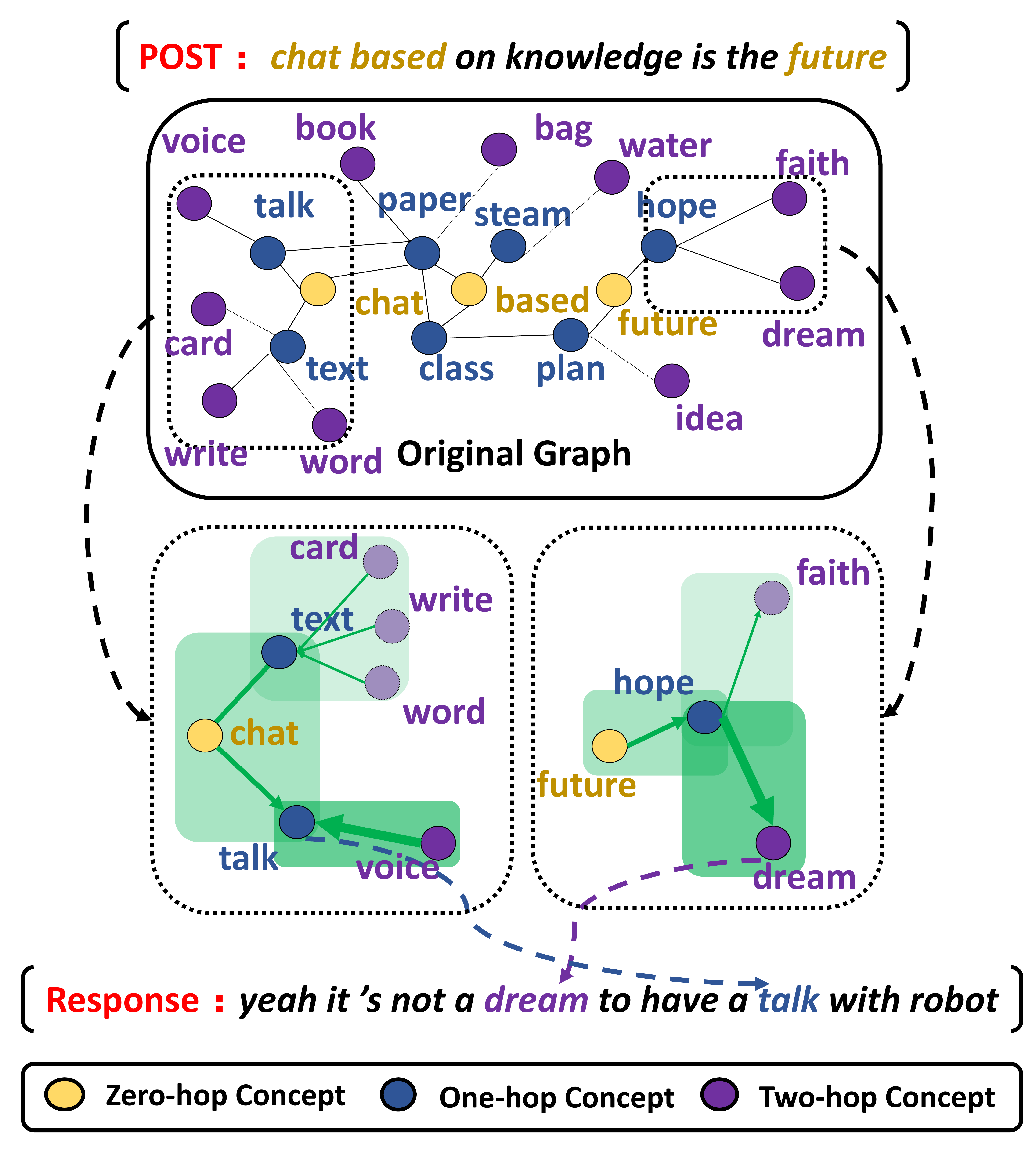}
 \caption{An Example of Concept Shift in a Conversation. Darker green indicates higher relevance and wider arrow indicates stronger concept shift (captured by ConceptFlow). \label{fig: motivation}}
\end{figure}

Conversations often develop around Knowledge. A promising way to address the degeneration problem is to ground conversations with external knowledge~\cite{xing2017topic}, such as open-domain knowledge graph~\cite{ghazvininejad2018knowledge}, commonsense knowledge base~\cite{zhou2018commonsense}, or background documents~\cite{zhou2018dataset}.
Recent research leverages such external knowledge by using them to ground conversations, integrating them as additional representations, and then generating responses conditioned on both the texts and the grounded semantics~\cite{ghazvininejad2018knowledge, zhou2018commonsense, zhou2018dataset}.

Integrating external knowledge as extra semantic representations and additional inputs to the conversation model effectively improves the quality of generated responses~\cite{ghazvininejad2018knowledge, logan2019barack, zhou2018commonsense}. Nevertheless, some research on discourse development suggests that human conversations are not ``still'': People chat around a number of related concepts, and shift their focus from one concept to others. ~\citet{grosz1986attention} models such concept shift by breaking discourse into several segments, and demonstrating different concepts, such as objects and properties, are needed to interpret different discourse segments. Attentional state is then introduced to represent the concept shift corresponding to each discourse segment. ~\citet{fang2018sounding} shows that people may switch dialog topics entirely in a conversation. Restricting the utilization of knowledge only to those directly appear in the conversation, effective as they are, does not reach the full potential of knowledge in modeling human conversations.


To model the concept shift in human conversations, this work presents ConceptFlow (\textbf{Con}versation generation with Con\textbf{cept} \textbf{Flow}), which leverages commonsense knowledge graphs to model the conversation flow in the explicit concept space.
For example, as shown in Figure~\ref{fig: motivation}, the concepts of a conversation from Reddit evolves from ``\textcolor[rgb]{0.5,0.5,0}{chat}'' and ``\textcolor[rgb]{0.5,0.5,0}{future}'', to adjacent concept ``\textcolor[rgb]{0,0,0.7}{talk}'', and also hops to distant concept ``\textcolor[rgb]{0.5,0,0.7}{dream}'' along the commonsense relations---a typical involvement in natural conversations. To better capture this conversation structure,
ConceptFlow explicitly models the conversations as traverses in commonsense knowledge graphs: it starts from the grounded concepts, e.g., ``\textcolor[rgb]{0.5,0.5,0}{chat}'' and ``\textcolor[rgb]{0.5,0.5,0}{future}'', and generates more meaningful conversations by hopping along the commonsense relations to related concepts, e.g., ``\textcolor[rgb]{0,0,0.7}{talk}'' and ``\textcolor[rgb]{0.5,0,0.7}{dream}''.


The traverses in the concept graph are guided by graph attention mechanisms, which derives from graph neural networks to attend on more appropriate concepts.
ConceptFlow learns to model the conversation development along more meaningful relations in the commonsense knowledge graph.
As a result, the model is able to ``grow'' the grounded concepts by hopping from the conversation utterances, along the commonsense relations, to distant but meaningful concepts; this guides the model to generate more informative and on-topic responses. Modeling commonsense knowledge as concept flows, is both a good practice on improving response diversity by scattering current conversation focuses to other concepts~\cite{chen2017survey}, and an implementation solution of the attentional state mentioned above~\cite{grosz1986attention}.


Our experiments on a Reddit conversation dataset with a commonsense knowledge graph, ConceptNet~\cite{speer2017conceptnet}, demonstrate the effectiveness of ConceptFlow.
In both automatic and human evaluations, ConceptFlow significantly outperforms various seq2seq based generation models~\cite{sutskever2014sequence}, as well as previous methods that also leverage commonsense knowledge graphs, but as static memories~\cite{zhou2018commonsense,ghazvininejad2018knowledge,zhu2017flexible}. Notably, ConceptFlow also outperforms two fine-tuned GPT-2 systems~\cite{radford2019language}, while using \textit{70\% fewer} parameters. Explicitly modeling conversation structure provides better parameter efficiency.

We also provide extensive analyses and case studies to investigate the advantage of modeling conversation flow in the concept space.
Our analyses show that many Reddit conversations are naturally aligned with the paths in the commonsense knowledge graph; incorporating distant concepts significantly improves the quality of generated responses with more on-topic semantic information added.
Our analyses further confirm the effectiveness of our graph attention mechanism in selecting useful concepts, and ConceptFlow's ability in leveraging them to generate more relevant, informative, and less repetitive responses.
\section{Related Work}
Sequence-to-sequence models, e.g., ~\citet{sutskever2014sequence}, have been widely used for natural language generation (NLG), and to build conversation systems~\citep{shang2015neural,vinyals2015neural,li2016deep,Wu2019Transferable}. Recently, pre-trained language models, such as ELMO~\citep{devlin2019bert}, UniLM~\citep{dong2019unified} and GPT-2~\citep{radford2016improving}, further boost the NLG performance with large scale pretraining.
Nevertheless, the degenerating of irrelevant, off-topic, and non-useful responses is still one of the main challenges in conversational generation~\citep{rosset2020leading, tang2019target, zhang2018generating, gao2019jointly}.

Recent work focuses on improving conversation generation with external knowledge, for example, incorporating additional texts~\citep{ghazvininejad2018knowledge,vougiouklis2016neural,xu2016incorporating,long2017knowledge}, or knowledge graphs~\citep{long2017knowledge, ghazvininejad2018knowledge}. They have shown external knowledge effectively improves conversation response generation.

The structured knowledge graphs include rich semantics represented via entities and relations~\citep{hayashi2019latent}.
Lots of previous studies focus on task-targeted dialog systems based on domain-specific knowledge bases~\citep{xu2016incorporating, zhu2017flexible, gu2016incorporating}. To generate responses with a large-scale knowledge base, \citet{zhou2018commonsense} and \citet{liu2018knowledge} utilize graph attention and knowledge diffusion to select knowledge semantics for utterance understanding and response generation. ~\citet{moon2019opendialkg} focuses on the task of entity selection, and takes advantage of positive entities that appear in the golden response. 
Different from previous research, ConceptFlow models the conversation flow explicitly with the commonsense knowledge graph and presents a novel attention mechanism on all concepts to guide the conversation flow in the latent concept space.

\section{Methodology}
This section presents our \textbf{Con}versation generation model with latent Con\textbf{cept} \textbf{Flow} (ConceptFlow). Our model grounds the conversation in the concept graph and traverses to distant concepts along commonsense relations to generate responses.

\subsection{Preliminary}
Given a user utterance $X = \{x_1, ..., x_m\}$ with $m$ words, conversation generation models often use an encoder-decoder architecture to generate a response $Y = \{y_1, ..., y_n\}$.

The encoder represents the user utterance $X$ as a representation set $H = \{\vec{h}_1, ..., \vec{h}_m\}$. This is often done by Gated Recurrent Units (GRU):
\begin{equation}
\small
 \vec{h}_i = \text{GRU} (\vec{h}_{i-1}, \vec{x}_i),
\end{equation}
where the $\vec{x}_i$ is the embedding of word $x_i$. 

The decoder generates $t$-th word in the response according to the previous $t-1$ generated words $y_{<t} = \{y_1,...,y_{t-1}\}$ and the user utterance $X$:
\begin{equation}
\small
 P(Y|X) = \prod^{n}_{t=1} P(y_t| y_{<t}, X). \label{eqn.prediction}
\end{equation}

Then it minimizes the cross-entropy loss $L$ and optimizes all parameters end-to-end:
\begin{equation}
\small
 L = \sum_{t=1}^{n}\text{CrossEntropy}(y_t^*, y_t), \label{eqn.loss}
\end{equation}
where $y_t^*$ is the token from the golden response.

\begin{figure}[t]
 \centering
 \includegraphics[width=1.0\linewidth]{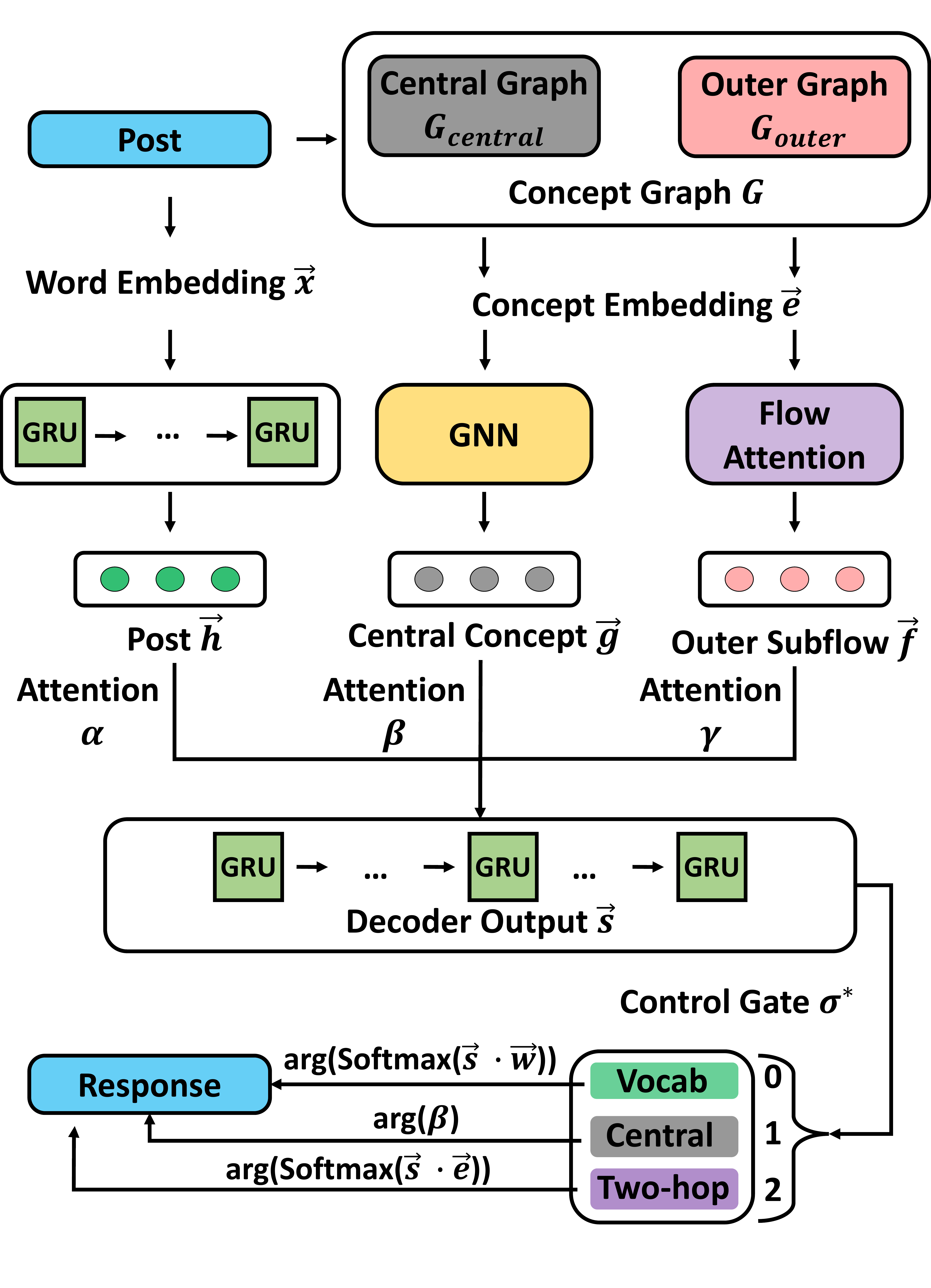}
 \caption{The Architecture of ConceptFlow. \label{fig:model}
 }

\end{figure}

The architecture of ConceptFlow is shown in Figure~\ref{fig:model}. ConceptFlow first constructs a concept graph $G$ with central graph $G_{\text{central}}$ and outer graph $G_{\text{outer}}$ according to the distance (hops) from the grounded concepts (Sec.~\ref{model:construction}).

Then ConceptFlow encodes both central and outer concept flows in central graph $G_{\text{central}}$ and outer graph $G_{\text{outer}}$ , using graph neural networks and concept embedding (Sec.~\ref{model:encoding}).

The decoder, presented in Section~\ref{model:generation}, leverages the encodings of concept flows and the utterance to generate words or concepts for responses.

\subsection{Concept Graph Construction}\label{model:construction}
ConceptFlow constructs a concept graph $G$ as the knowledge for each conversation. It starts from the grounded concepts (zero-hop concepts $V^0$), which appear in the conversation utterance and annotated by entity linking systems.

Then, ConceptFlow grows zero-hop concepts $V^0$ with one-hop concepts $V^1$ and two-hop concepts $V^2$. Concepts from $V^0$ and $V^1$, as well as all relations between them, form the central concept graph $G_\text{central}$, which is closely related to the current conversation topic. Concepts in $V^1$ and $V^2$ and their connections form the outer graph $G_\text{outer}$. 

\subsection{Encoding Latent Concept Flow}\label{model:encoding}
The constructed concept graph provides explicit semantics on how concepts related to commonsense knowledge. ConceptFlow utilizes it to model the conversation and guide the response generation. It starts from the user utterance, traversing through central graph $G_\text{central}$, to outer graph $G_\text{outer}$.
This is modeled by encoding the central and outer concept flows according to the user utterance.

\textbf{Central Flow Encoding.} The central concept graph  $G_\text{central}$ is encoded by a graph neural network that propagates information from user utterance $H$ to the central concept graph. Specifically, it encodes concept $e_i \in G_\text{central}$ to representation $\vec{g}_{e_i}$:
\begin{equation}
\small
\vec{g}_{e_i} = \text{GNN} (\vec{e}_i, G_\text{central}, H),
\end{equation}
where $\vec{e}_i$ is the concept embedding of $e_i$. 
There is no restriction of which GNN model to use. We choose~\citet{sun2018open}'s GNN (GraftNet), which shows strong effectiveness in encoding knowledge graphs. More details of GraftNet can be found in Appendix \ref{appendix model}.

\textbf{Outer Flow Encoding.}
The outer flow $f_{e_p}$, hopping from $e_p \in V_1$ to its connected two-hop concept $e_k$, is encoded to $\vec{f}_{e_p}$ by an attention mechanism:
\begin{equation}
\small
\vec{f}_{e_p} =\sum_{e_k} \theta^{e_k} \cdot [ \vec{e}_p \circ \vec{e}_k ], 
\end{equation}
where $\vec{e}_p$ and $\vec{e}_k$ are embeddings for $e_p$ and $e_k$, and are concatenated ($\circ$). The attention $\theta^{e_k}$ aggregates concept triple $(e_p, r, e_k)$ to get $\vec{f}_{e_p}$:
\begin{equation}
\small
\theta^{e_k} = \text{softmax} ((w_r \cdot \vec{r})^{\top} \cdot \tanh (w_h \cdot \vec{e}_p+ w_t \cdot \vec{e}_k)),
\end{equation}
where $ \vec{r}$ is the relation embedding between the concept $e_p$ and its neighbor concept $e_k$. $w_r$, $w_h$ and $w_t$ are trainable parameters. It provides an efficient attention specifically focusing on the relations for multi-hop concepts.

\subsection{Generating Text with Concept Flow}\label{model:generation}
To consider both user utterance and related information, 
the texts from the user utterance and the latent concept flows are incorporated by decoder using two components: 1) the context representation that combines their encodings (Sec.~\ref{sec:contextrep}); 2) the conditioned generation of words and concepts from the context representations (Sec.~\ref{sec:genertingtokens}).

\subsubsection{Context Representation}\label{sec:contextrep}
To generate $t$-th time response token, we first calculate the output context representation $\vec{s}_t$ for $t$-th time decoding with the encodings of the utterance and the latent concept flow.

Specifically, $\vec{s}_t$ is calculated by updating the ($t-1$)-th step output representation $\vec{s}_{t-1}$ with the ($t-1$)-th step context representation $\vec{c}_{t-1}$:
\begin{equation}\label{equ:st}
\small
\vec{s}_{t}=\text{GRU}(\vec{s}_{t-1},[\vec{c}_{t-1} \circ \vec{y}_{t-1}]),
\end{equation}
where $\vec{y}_{t-1}$ is the ($t-1$)-th step generated token $y_{t-1}$'s embedding, and the context representation $\vec{c}_{t-1}$ concatenates the text-based representation $\vec{c}_{t-1}^{\text{\, text}}$ and the concept-based representation $\vec{c}_{t-1}^{\text{\, concept}}$:
\begin{equation}
\small
\vec{c}_{t-1} = \text{FFN}([\vec{c}_{t-1}^{\text{\, text}} \circ \vec{c}_{t-1}^{\text{\, cpt}}]).
\end{equation}

The \textbf{text-based representation} $\vec{c}_{t-1}^{\text{\, text}}$ reads the user utterance encoding $H$ with a standard attention mechanism~\cite{bahdanau2014neural}:
\begin{equation}
\small
\begin{aligned}
 \vec{c}_{t-1}^{\text{\, text}} &= \sum_{i=1}^{m} \alpha^{j}_{t-1} \cdot \vec{h}_j,
\end{aligned}
\end{equation}
and attentions $\alpha_{t-1}^{j}$ on the utterance tokens:
\begin{equation}
\small
\alpha_{t-1}^{j} =\text{softmax}(\vec{s}_{t-1} \cdot \vec{h}_j).
\end{equation}

The \textbf{concept-based representation} $\vec{c}_{t-1}^{\text{\, concept}}$ is a combination of central and outer flow encodings:
\begin{equation}
\small
 \vec{c}_{t-1}^{\text{\, cpt}}= \left( \sum_{e_i \in G_\text{central}} \beta^{e_i}_{t-1} \cdot \vec{g}_{e_i} \right)\circ \left( \sum_{f_{e_p} \in G_{\text{outer}}} \gamma^{f}_{t-1} \cdot \vec{f}_{e_p} \right).
\end{equation}
The attention $\beta_{t-1}^{e_i}$ weights over central concept representations:
\begin{equation}
\small
\beta_{t-1}^{e_i} =\text{softmax}(\vec{s}_{t-1} \cdot \vec{g}_{e_i}),
\end{equation}
and the attention $\gamma_{t-1}^{f}$ weights over outer flow representations:
\begin{equation}
\small
\gamma_{t-1}^{f} =\text{softmax}(\vec{s}_{t-1} \cdot \vec{f}_{e_p}).
\end{equation}

\subsubsection{Generating Tokens}\label{sec:genertingtokens}
The $t$-th time output representation $\vec{s}_t$ (Eq.~\ref{equ:st}) includes information from both the utterance text, the concepts with different hop steps, and the attentions upon them.
The decoder leverages $\vec{s}_t$ to generate the $t$-th token to form more informative responses.

It first uses a gate $\sigma^{*}$ to control the generation by choosing words ($\sigma^{*}=0$), central concepts ($V^{0,1}$, $\sigma^{*}=1$) and outer concept set ($V^{2}$, $\sigma^{*}=2$):
\begin{equation}
\small
 \sigma^{*} = \text{argmax}_{\sigma \in \{0, 1, 2\}}(\text{FFN}_{\sigma}(\vec{s}_t)),
\end{equation}

The generation probabilities of word $w$, central concept $e_i$, and outer concepts $e_k$ are calculated over the word vocabulary, central concept set $V^{0,1}$, and outer concept set $V^{2}$:
\begin{equation}
\small
y_t \sim \left\{\begin{matrix}\begin{aligned}
&\text{softmax}(\vec{s}_{t} \cdot \vec{w}), \sigma^{*} = 0 \\ 
&\text{softmax}(\vec{s}_{t} \cdot \vec{g}_{e_i}), \sigma^{*} = 1 \\
&\text{softmax}(\vec{s}_{t} \cdot \vec{e}_k), \sigma^{*} = 2, \\\end{aligned}
\end{matrix}\right. \label{eqn.finaly}
\end{equation}
where $\vec{w}$ is the word embedding for word $w$, $\vec{g}_{e_i}$ is the central concept representation for concept $e_i$ and $\vec{e}_k$ is the two-hop concept $e_k$'s embedding.

The training and prediction of ConceptFlow are conducted following standard conditional language models, i.e. using Eq.~\ref{eqn.finaly} in place of Eq.~\ref{eqn.prediction} and training it by the Cross-Entropy loss  (Eq.~\ref{eqn.loss}). Only ground truth responses are used in training and no additional annotation is required.

\section{Experiment Methodology}
This section describes the dataset, evaluation metrics, baselines, and implementation details of our experiments.

\textbf{Dataset.}
All experiments use the multi-hop extended conversation dataset based on a previous dataset which collects single-round dialogs from Reddit~\cite{zhou2018commonsense}. Our dataset contains 3,384,185 training pairs and 10,000 test pairs. Preprocessed ConceptNet~\cite{speer2017conceptnet} is used as the knowledge graph, which contains 120,850 triples, 21,471 concepts and 44 relation types.

\textbf{Evaluation Metrics.}
A wide range of evaluation metrics are used to evaluate the quality of generated responses: PPL~\cite{serban2016building}, Bleu~\cite{papineni2002bleu}, Nist~\cite{doddington2002automatic}, ROUGE~\cite{lin2004rouge} and Meteor~\cite{lavie2007meteor} are used for relevance and repetitiveness; Dist-1, Dist-2 and Ent-4 are used for diversity, which is same with the previous work~\cite{li2015diversity,zhang2018generating}. The metrics above are evaluated using the implementation from~\citet{galley2018end}. \citet{zhou2018commonsense}'s concept PPL mainly focuses on concept grounded models and this metric is reported in Appendix~\ref{appendix A.1}.

The Precision, Recall, and F1 scores are used to evaluate the quality of learned latent concept flow in predicting the golden concepts which appear in ground truth responses.

\textbf{Baselines.} The six baselines compared come from three groups: standard Seq2Seq, knowledge-enhanced ones, and fine-tuned GPT-2 systems.

Seq2Seq~\cite{sutskever2014sequence} is the basic encoder-decoder for language generation.  

Knowledge-enhanced baselines include
MemNet~\cite{ghazvininejad2018knowledge}, CopyNet~\cite{zhu2017flexible} and CCM~\cite{zhou2018commonsense}.
MemNet maintains a memory to store and read concepts.
CopyNet copies concepts for the response generation. 
CCM~\cite{zhou2018commonsense} leverages a graph attention mechanism to model the central concepts. These models mainly focus on the grounded concepts. They do not explicitly model the conversation structures using multi-hop concepts.

GPT-2~\cite{radford2019language}, the pre-trained model that achieves the state-of-the-art in lots of language generation tasks, is also compared in our experiments. We fine-tune the 124M GPT-2 in two ways: concatenate all conversations together and train it like a language model (GPT-2 \textit{lang}); extend the GPT-2 model with encode-decoder architecture and supervise with response data (GPT-2 \textit{conv}).

\textbf{Implement Details.}
The zero-hop concepts are initialized by matching the keywords in the post to concepts in ConceptNet, the same with CCM~\cite{zhou2018commonsense}. Then zero-hop concepts are extended to their neighbors to form the central concept graph. The outer concepts contain a large amount of two-hop concepts with lots of noises. To reduce the computational cost, we first train ConceptFlow (select) with 10\% random training data, and use the learned graph attention to select top 100 two-hop concepts over the whole dataset. Then the standard train and test are conducted with the pruned graph. More details of this filtering step can be found in Appendix~\ref{appendix C}.

TransE~\cite{bordes2013translating} embedding and Glove~\cite{pennington2014glove} embedding are used to initialize the representation of concepts and words, respectively. Adam optimizer with the learning rate of 0.0001 is used to train the model.
\begin{table*}[t]
\centering
\small
\begin{tabular}{l|c|c|c|c|c|c|c}
\hline  \hline
Model     & Bleu-4 & Nist-4 & Rouge-1 & Rouge-2 & Rouge-L & Meteor & PPL \\ \hline
Seq2Seq   & 0.0098 & 1.1069 & 0.1441  & 0.0189  & 0.1146  & 0.0611 & 48.79 \\ \hline
MemNet    & 0.0112 & 1.1977 & 0.1523  & 0.0215  & 0.1213  & 0.0632 & 47.38 \\ 
CopyNet   & 0.0106 & 1.0788 & 0.1472  & 0.0211  & 0.1153  & 0.0610 & 43.28 \\
CCM       & 0.0084 & 0.9095 & 0.1538  & 0.0211  & 0.1245  & 0.0630 & 42.91 \\ \hline
GPT-2 (lang) & 0.0162 & 1.0844 & 0.1321 & 0.0117 & 0.1046 & 0.0637 & 29.08$^*$ \\ 
GPT-2 (conv)     & 0.0124 & 1.1763 & 0.1514  & 0.0222  & 0.1212  & 0.0629 & 24.55$^*$ \\ \hline
ConceptFlow & \textbf{0.0246} & \textbf{1.8329} & \textbf{0.2280}  & \textbf{0.0469}  & \textbf{0.1888}  & \textbf{0.0942} & \textbf{29.90} \\ \hline  \hline
\end{tabular}
\caption{\label{tab: final relevance}Relevance Between Generated and Golden Responses. The PPL results$^*$ of GPT-2 is not directly comparable because of its different tokenization. More results can be found in Appendix~\ref{appendix A.1}.
}
\end{table*}
\begin{table*}[t]
\centering
\small
\begin{tabular}{l|c|c|c||c|c|c|c|c}
\hline  \hline
& \multicolumn{3}{c||}{Diversity($\uparrow$)} & \multicolumn{5}{c}{Novelty w.r.t. Input($\downarrow$)} \\ \hline
Model     &Dist-1 & Dist-2 & Ent-4 & Bleu-4 & Nist-4 & Rouge-2 & Rouge-L & Meteor  \\ \hline
Seq2Seq   & 0.0123 & 0.0525 & 7.665 & 0.0129 & 1.3339 & 0.0262 & \textbf{0.1328}  & \textbf{0.0702}   \\ \hline
MemNet    & 0.0211 & 0.0931 & 8.418 & 0.0408 & 2.0348 & 0.0621 & 0.1785  & 0.0914 \\ 
CopyNet   & 0.0223 & 0.0988 & 8.422 & 0.0341 & 1.8088 & 0.0548 & 0.1653  & 0.0873   \\ 
CCM      & 0.0146 & 0.0643 & 7.847 & 0.0218 & \textbf{1.3127} & 0.0424 & 0.1581  & 0.0813 \\ \hline
GPT-2 (lang) & \textbf{0.0325} & \textbf{0.2461} & \textbf{11.65} & 0.0292 & 1.7461 & 0.0359 & 0.1436 & 0.0877 \\ 
GPT-2 (conv)     & 0.0266 & 0.1218 & 8.546 & 0.0789 & 2.5493 & 0.0938 & 0.2093  & 0.1080   \\ \hline
ConceptFlow & 0.0223 & 0.1228 & 10.27 & \textbf{0.0126} & 1.4749 & \textbf{0.0258} & 0.1386  & 0.0761  \\ \hline \hline 
\end{tabular}
\caption{\label{tab: final diversity}Diversity (higher better) and Novelty (lower better) of Generated Response. Diversity is calculated within generated responses;  Novelty compares generated responses to the input post.
More results are in Appendix~\ref{appendix A.1}.}
\end{table*}
\begin{table}[t]
\centering
\small
\resizebox{0.49\textwidth}{!}{
\begin{tabular}{l|c|c|c|c|c}
\hline \hline
\multirow{2}{*}{Model}& \multirow{2}{*}{Parameter}& \multicolumn{2}{c|}{Average Score} & \multicolumn{2}{c}{Best@1 Ratio} \\ 
\cline{3-6}
& & App. & Inf. & App. & Inf. \\ \hline
CCM   &35.6M& 1.802 & 1.802 & 17.0\% & 15.6\%     \\ \hline
GPT-2 (conv) &124.0M& 2.100 & 1.992 & 26.2\%  & 23.6\%     \\ \hline
ConceptFlow   &35.3M& 2.690 & 2.192 & 30.4\% & 25.6\%     \\ \hline
Golden  & Human & 2.902 & 3.110 & 67.4\% & 81.8\%       \\ \hline \hline
\end{tabular}}
\caption{\label{tab: human_evaluation}Human Evaluation on Appropriate (App.) and Informativeness (Inf.). The Average Score takes the average from human judgments. Best@1 Ratio indicates the fraction of judges consider the case as the best. The number of parameters are also presented.}
\end{table}
\begin{table}[t]
\centering
\begin{tabular}{l|c|c}
\hline  \hline
Model     & App. & Inf. \\ \hline
ConceptFlow-CCM   & 0.3724 & 0.2641 \\ \hline
ConceptFlow-GPT2  & 0.2468  & 0.2824 \\ \hline  \hline
\end{tabular}
\caption{\label{tab: kappa}Fleiss' Kappa of Human Agreement. Two testing scenarios Appropriate (App.) and Informativeness (Inf.) are used to evaluate the the quality of generated response. The Fleiss' Kappa evaluates agreement from various annotators and focuses on the comparison of two models with three categories: win, tie and loss.}
\end{table}
\section{Evaluation}
Five experiments are conducted to evaluate the generated responses from ConceptFlow and the effectiveness of the learned graph attention.

\subsection{Response Quality}
This experiment evaluates the generation quality of ConceptFlow automatically and manually.

\textbf{Automatic Evaluation.}
The quality of generated responses is evaluated with different metrics from three aspects: relevance, diversity, and novelty. Table~\ref{tab: final relevance} and Table~\ref{tab: final diversity} show the results.

In Table~\ref{tab: final relevance}, all evaluation metrics calculate the relevance between the generated response and the golden response. ConceptFlow outperforms all baseline models by large margins. The responses generated by ConceptFlow are more on-topic and match better with the ground truth responses.

In Table~\ref{tab: final diversity}, Dist-1, Dist-2, and Ent-4 measure the word diversity of generated responses and the rest of metrics measure the novelty by comparing the generated response with the user utterance. ConceptFlow has a good balance in generating novel and diverse responses. GPT-2's responses are more diverse, perhaps due to its sampling mechanism during decoding, but are less novel and on-topic compared to those from ConceptFlow.

\textbf{Human Evaluation.}
The human evaluation focuses on two aspects: appropriateness and informativeness. Both are important for conversation systems~\cite{zhou2018commonsense}. Appropriateness evaluates if the response is on-topic for the given utterance; informativeness evaluates systems' ability to provide new information instead of copying from the utterance~\cite{zhou2018commonsense}. All responses of sampled 100 cases are selected from four methods with better performances: CCM, GPT-2 (conv), ConceptFlow, and Golden Response. The responses are scored from 1 to 4 by five judges (the higher the better). 

Table~\ref{tab: human_evaluation} presents Average Score and Best@1 ratio from human judges. The first is the mean of five judges; the latter calculates the fraction of judges that consider the corresponding response the best among four systems. ConceptFlow outperforms all other models in all scenarios, while only using 30\% of parameters compared to GPT-2.
This demonstrates the advantage of explicitly modeling conversation flow with structured semantics. 

The agreement of human evaluation is tested to demonstrate the authenticity of evaluation results.
We first sample 100 cases randomly for our human evaluation. Then the responses from four better conversation systems, CCM, GPT-2 (conv), ConceptFlow and Golden Responses, are provided with a random order. A group of annotators are asked to score each response ranged from 1 to 4 according to the quality on two testing scenarios, appropriateness and informativeness. All annotators have no clues about the source of generated responses.

The agreement of human evaluation for CCM, GPT-2 (conv) and ConceptFlow are presented in Table~\ref{tab: kappa}. For each case, the response from ConceptFlow is compared to the responses from two baseline models, CCM and GPT-2 (conv). The comparison result is divided into three categories: win, tie and loss. Then the human evaluation agreement is calculated with Fleiss' Kappa ($\kappa$).  The $\kappa$ value ranges from 0.21 to 0.40 indicating fair agreement, which confirms the quality of human evaluation.

Both automatic and human evaluations illustrate the effectiveness of ConceptFlow. The next experiment further studies the effectiveness of multi-hop concepts in ConceptFlow.

\begin{figure*}[t]
\centering   
    \subfigure[Golden Concept Coverage.]{\label{fig:ratio}\includegraphics[width=0.325\linewidth]{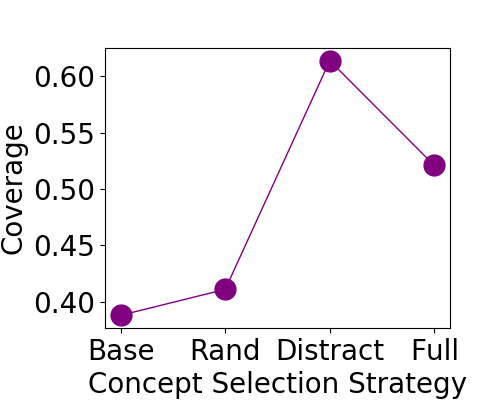}} 
    \subfigure[Response Concept Generation.]{\label{fig:pr}\includegraphics[width=0.325\linewidth]{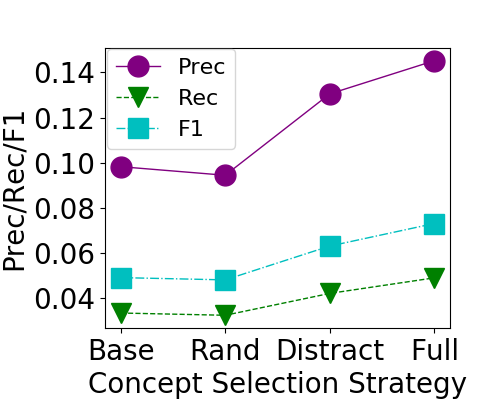}} 
    \subfigure[Response Token Generation.]{\label{fig:ppl}\includegraphics[width=0.325\linewidth]{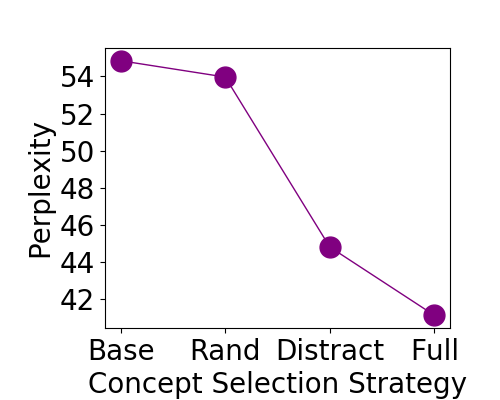}}
    \caption{Comparisons of Outer Concept Selection Methods. \textit{Base} only considers the central concepts and ignores two-hop concepts. \textit{Rand} randomly selects two-hop concepts. \textit{Distract} incorporates golden concepts in the response with random negatives (distractors). \textit{Full} chooses two-hop concepts with ConceptFlow's graph attention.}
    \label{fig:ablation}
\end{figure*}
\begin{figure*}[t]
\centering
    \includegraphics[width=0.9\linewidth]{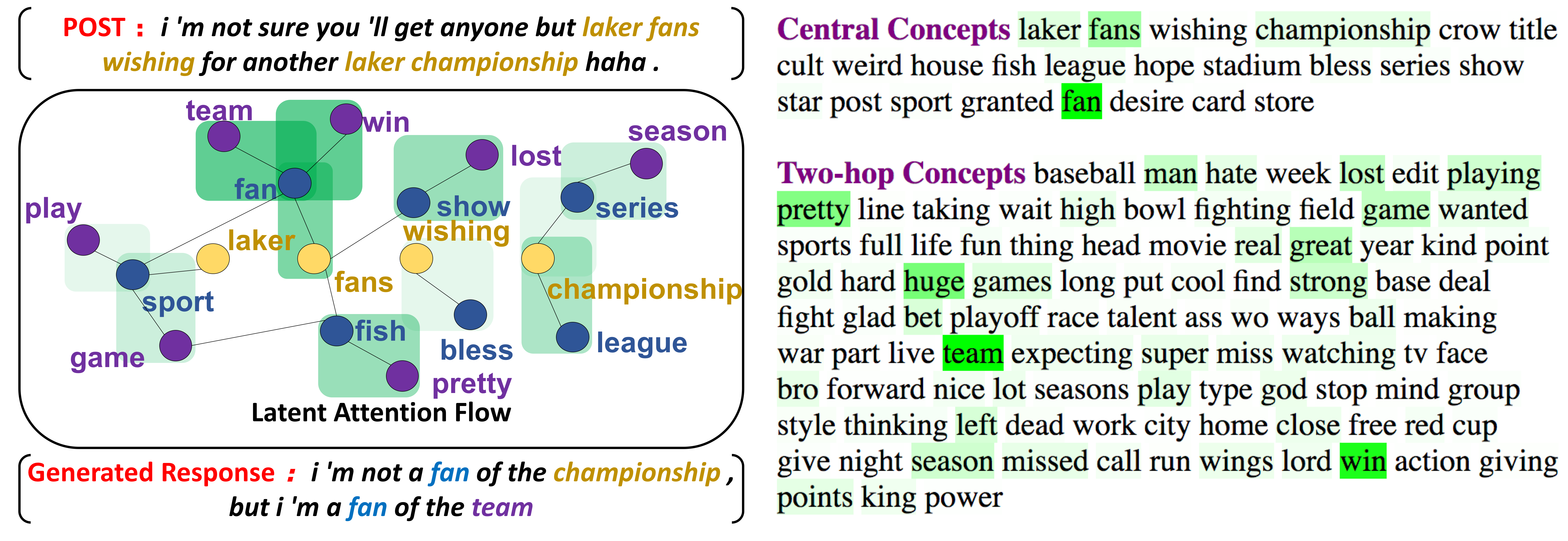}
    \caption{Case Study (Best viewed in color). \textbf{Left:} Attention flow in commonsense concept graph, where \textcolor[rgb]{0.5,0.5,0}{zero-hop concepts}, \textcolor[rgb]{0,0,0.7}{one-hop concepts} and \textcolor[rgb]{0.5,0,0.7}{two-hop concepts} are highlighted. \textbf{Right:} Attention scores over all concepts. Darker green indicates higher attention scores. }
    \label{fig:attention}
\end{figure*}
\begin{figure}[t]
\centering   
    \subfigure[Central Concept.]{\label{fig:central}\includegraphics[width=0.49\linewidth]{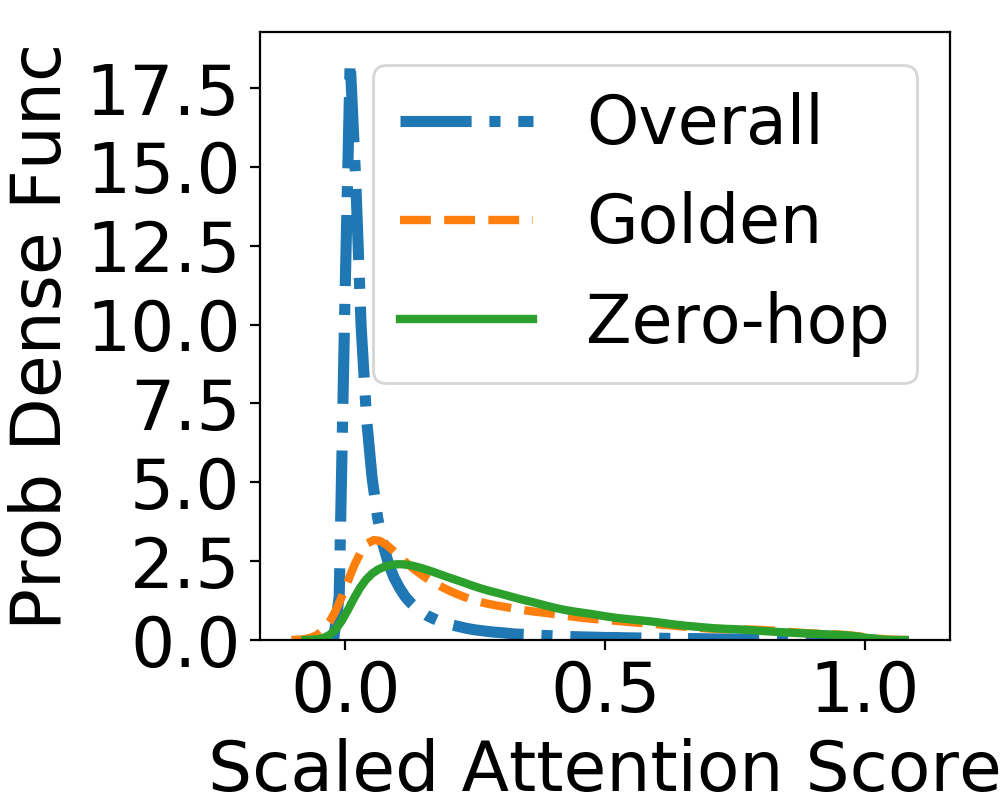}} 
    \subfigure[Two-hop Concept.]{\label{fig:outer}\includegraphics[width=0.49\linewidth]{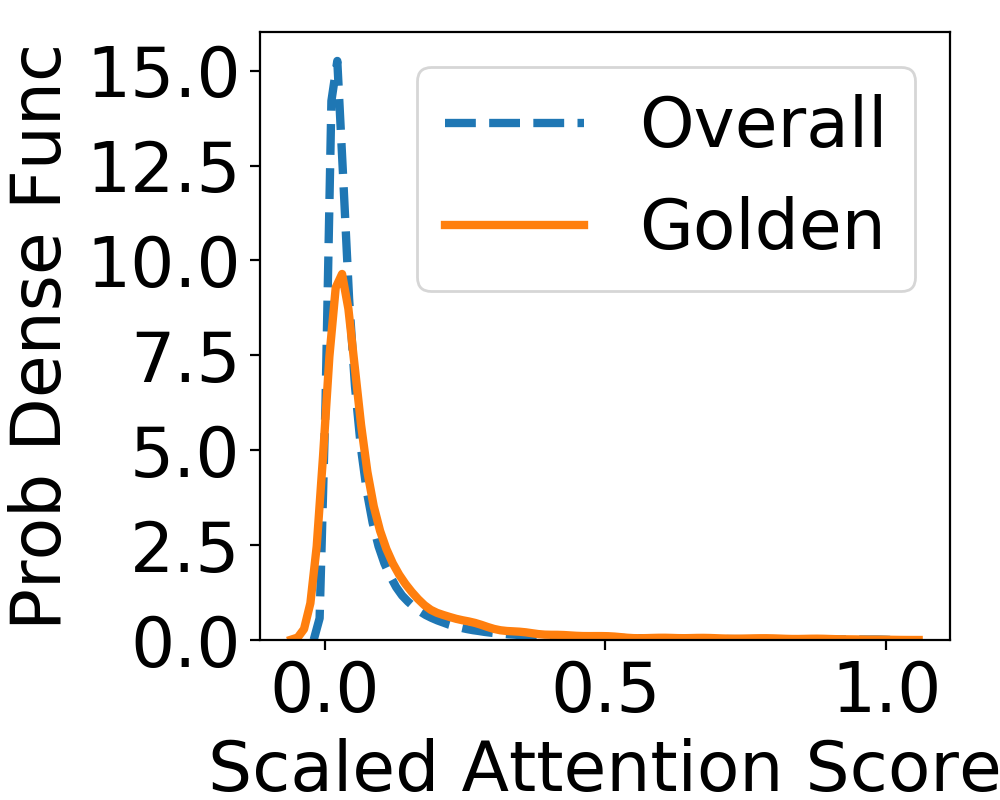}} 
    \caption{Distribution of Attention Score. The distributions of Overall (all concepts of the certain part), Golden (concepts in the golden response) and Zero-hop (concepts appear in the post) are presented. The attention score is calculated by scaling the mean of attention scores of $n$ step decoding.}
    \label{fig:attention distribution}
\end{figure}

\subsection{Effectiveness of Multi-hop Concepts} \label{evaluation 2}
This part explores the role of multi-hop concepts in ConceptFlow. As shown in Figure~\ref{fig:ablation}, three experiments are conducted to evaluate the performances of concept selection and the quality of generated responses with different sets of concepts. 

This experiment considers four variations of outer concept selections. \textit{Base} ignores two-hop concepts and only considers the central concepts. \textit{Rand}, \textit{Distract}, and \textit{Full} add two-hop concepts in three different ways: \textit{Rand} selects concepts randomly, \textit{Distract} selects all concepts that appear in the golden response with random negatives (distractors), and \textit{Full} is our ConceptFlow (select) that selects concepts by learned graph attentions.

As shown in Figure~\ref{fig:ratio}, \textit{Full} covers more golden concepts than \textit{Base}. This aligns with our motivation that natural conversations do flow from central concepts to multi-hop ones. 
Compared to \textit{Distract} setting where all ground truth two-hop concepts are added, ConceptFlow (select) has slightly less coverage but significantly reduces the number of two-hop concepts. 

The second experiment studies the model's ability to generate ground truth concepts, by comparing the concepts in generated responses with those in ground truth responses.
As shown in Figure~\ref{fig:pr}, though \textit{Full} filtered out some golden two-hop concepts, it outperforms other variations by large margins. This shows ConceptFlow's graph attention mechanisms effectively leverage the pruned concept graph and generate high-quality concepts when decoding.


The high-quality latent concept flow leads to better modeling of conversations, as shown in Figure~\ref{fig:ppl}. \textit{Full} outperforms \textit{Distract} in their generated responses' token level perplexity, even though \textit{Distract} includes all ground truth two-hop concepts. This shows that ``negatives'' selected by ConceptFlow, while not directly appear in the target response, are also on-topic and include meaningful information, as they are selected by graph attentions instead of random.

More studies of multi-hop concept selection strategies can be found in Appendix~\ref{appendix A.2}.

\begin{table}[t]
\centering
\small
\begin{tabular}{l|c|c|c}
\hline  \hline
\multirow{2}{*}{Depth}     & \multirow{2}{*}{Amount} & \multicolumn{2}{c}{Golden Coverage}  \\ \cline{3-4}
& & Ratio & Number \\ \hline
Zero-hop   & 5.8 & 9.81\% & 0.579      \\ 
+ One-hop    & 98.6 & 38.78\% & 2.292 \\ 
+ Two-hop    & 880.8 & 61.37\% & 3.627 \\
+ Three-hop  & 3769.1 & 81.58\% & 4.821 \\ \hline
ConceptFlow  & 198.6 & 52.10\% & 3.075 \\ \hline
\hline
\end{tabular}
\caption{Statistics of Concept Graphs with different hops, including the total Amount of connected concepts, the Ratio and Number of covered golden concepts (those appear in ground truth responses). ConceptFlow indicates the filtered two-hop graph.
\label{tab: data_study}}
\end{table}
\begin{table*}[h]
\centering
\small
\resizebox{0.99\textwidth}{!}{
\begin{tabular}{l|l}
\hline \hline
\textbf{\#1 Post}          & \begin{tabular}[c]{@{}l@{}}\textbf{actually i \textcolor[rgb]{0.5,0.5,0}{stayed} at the \textcolor[rgb]{0.5,0.5,0}{building} right next to the \textcolor[rgb]{0.5,0.5,0}{lighthouse} . i believe it was}  \textbf{a \textcolor[rgb]{0.5,0.5,0}{boiler} \textcolor[rgb]{0.5,0.5,0}{room} .}\\ \textbf{another \textcolor[rgb]{0.5,0.5,0}{group} was already \textcolor[rgb]{0.5,0.5,0}{sleeping} at the \textcolor[rgb]{0.5,0.5,0}{lighthouse}} .\end{tabular} \\ \hline
CCM           & \begin{tabular}[c]{@{}l@{}}i 'm not sure if you 're joking , but i 'm not sure if you 're talking about the lighthouse . \\i 'm not sure if it was a \textcolor[rgb]{0,0,0.7}{blanket} or a \textcolor[rgb]{0,0,0.7}{blanket} .\end{tabular}           \\
GPT-2 (conv) & \begin{tabular}[c]{@{}l@{}}i 'm pretty sure it was a room with a door .  \end{tabular}                                                                                                                                                     \\
ConceptFlow     & good \textcolor[rgb]{0.5,0,0.7}{luck} , i 'm sure there are some good things in the \textcolor[rgb]{0,0,0.7}{house} .                                                                                                                                     \\ 
 
\hline
\textbf{\#2 Post}          & \begin{tabular}[c]{@{}l@{}}\textbf{at this \textcolor[rgb]{0.5,0.5,0}{point} it may be more \textcolor[rgb]{0.5,0.5,0}{beneficial} for him to keep \textcolor[rgb]{0.5,0.5,0}{playing} as much as he}  \textbf{can before the next \textcolor[rgb]{0.5,0.5,0}{season} .}\\ \textbf{\textcolor[rgb]{0.5,0.5,0}{trick} his \textcolor[rgb]{0.5,0.5,0}{body} into \textcolor[rgb]{0.5,0.5,0}{thinking} he is still in the} \textbf{\textcolor[rgb]{0.5,0.5,0}{season} .}\end{tabular} \\ \hline
CCM           & i think he 's just going to be a \textcolor[rgb]{0,0,0.7}{great} player .                                                                                                                                                               \\ 
GPT-2 (conv) & i 'm sure he 'll be fine . he 's a great player , but he 's not a top 10 player .                                                                                                                             \\ 
ConceptFlow     & \begin{tabular}[c]{@{}l@{}}he 's been \textcolor[rgb]{0.5,0.5,0}{playing} the same \textcolor[rgb]{0,0,0.7}{game} for the last two \textcolor[rgb]{0.5,0,0.7}{years} . he 's been \textcolor[rgb]{0.5,0.5,0}{playing} the same way .\end{tabular}                                                                                                                                     \\ 
\hline
\textbf{\#3 Post}          & \begin{tabular}[c]{@{}l@{}}\textbf{i 've known how to \textcolor[rgb]{0.5,0.5,0}{change} the \textcolor[rgb]{0.5,0.5,0}{music} , it does n't \textcolor[rgb]{0.5,0.5,0}{feel} \textcolor[rgb]{0.5,0.5,0}{natural} . thanks for} \textbf{the \textcolor[rgb]{0.5,0.5,0}{advice} though .}\end{tabular}                                                                                                                                         \\ \hline
CCM           & \begin{tabular}[c]{@{}l@{}}i 'm not sure if you 're joking , but i 'm not sure if you 're being sarcastic or not . i 'm not sure if you 're\\ being sarcastic or not , but i 'm not sure if you 're being sarcastic or not . i 'm \textcolor[rgb]{0,0,0.7}{pretty} sure it 's just a video .\end{tabular} \\ 
GPT-2 (conv) & i 've been playing with a lot of different music and it 's always been the same .                                                                                                                                                                                                 \\ 
ConceptFlow     & i did n't even notice the \textcolor[rgb]{0.5,0.5,0}{natural} \textcolor[rgb]{0,0,0.7}{sound} . i 'll \textcolor[rgb]{0.5,0,0.7}{check} it out !                                                                                                                                     \\                            
\hline
\textbf{\#4 Post}          & \begin{tabular}[c]{@{}l@{}}\textbf{are \textcolor[rgb]{0.5,0.5,0}{milan} and \textcolor[rgb]{0.5,0.5,0}{inter} no \textcolor[rgb]{0.5,0.5,0}{longer} \textcolor[rgb]{0.5,0.5,0}{top} \textcolor[rgb]{0.5,0.5,0}{clubs} given that they currently do n't} \textbf{have the \textcolor[rgb]{0.5,0.5,0}{finances}} \\ \textbf{to \textcolor[rgb]{0.5,0.5,0}{compete} with the best ?}\end{tabular} \\ \hline
CCM           & \begin{tabular}[c]{@{}l@{}}milan is n't a top \textcolor[rgb]{0,0,0.7}{club} , but they are n't top clubs . they are n't the only ones who are n't in the \textcolor[rgb]{0,0,0.7}{world} .\end{tabular}          \\ 
GPT-2 (conv) & i think they are . they are a club that has been in the top 4 for a long time .                                                                                 \\ 
ConceptFlow     & \begin{tabular}[c]{@{}l@{}}the best \textcolor[rgb]{0,0,0.7}{part} of the \textcolor[rgb]{0.5,0,0.7}{team} is the way they are , and we do n't want to play the best \textcolor[rgb]{0.5,0,0.7}{team} in the \textcolor[rgb]{0.5,0,0.7}{country} .\end{tabular}                                                                                                                                     \\ 
\hline
\textbf{\#5 Post}          & \textbf{no i 've never \textcolor[rgb]{0.5,0.5,0}{felt} like i had \textcolor[rgb]{0.5,0.5,0}{friends} \textcolor[rgb]{0.5,0.5,0}{whoops}}                                                                                           \\ \hline
CCM           & i 'm sorry , i \textcolor[rgb]{0,0,0.7}{feel} you .                                                                                                               \\ 
GPT-2 (conv) & \begin{tabular}[c]{@{}l@{}}i 've never felt like i had friends whoops . i 've never felt like i 've had friends whoops .\end{tabular} \\ 
ConceptFlow     & you 're gonna have to \textcolor[rgb]{0,0,0.7}{call} your \textcolor[rgb]{0,0,0.7}{friend} !                                                                                                                                     \\

\hline\hline                                                                                                 
\end{tabular}}
\caption{\label{tab: case in appendix}Some Cases Generated by Different Models. Tokens from \textcolor[rgb]{0.5,0.5,0}{zero concepts}, \textcolor[rgb]{0,0,0.7}{one-hop concepts} and \textcolor[rgb]{0.5,0,0.7}{two-hop concepts} are highlighted.}

\end{table*}




\subsection{Hop Steps in Concept Graph}
This experiment studies the influence of hop steps in the concept graph.

As shown in Table~\ref{tab: data_study}, the Number of covered golden concepts increases with more hops. Compared to zero-hop concepts, multi-hop concepts cover more golden concepts, confirming that conversations naturally shift to multi-hop concepts: extending the concept graph from one-hop to two-hop improves the recall from 39\% to 61\%, and to three-hop further improves to 81\%.

However, at the same time, the amounts of the concepts also increase dramatically with multiple hops. Three hops lead to 3,769 concepts on average, which are 10\% of the entire graph we used. In this work, we choose two-hop, as a good balance of coverage and efficiency, and used ConceptFlow (select) to filter around 200 concepts to construct the pruned graph. How to efficiently and effectively leverage more distant concepts in the graph is reserved for future work.


\subsection{Case Study}
Some cases from three conversation models are listed in Table~\ref{tab: case in appendix}.
Responses from CCM may repeat the same contents as it does not explicitly model the traverse in the concept space.
For example, the responses from the first and third cases always repeat ``I'm not sure''.  On the other hand, GPT-2 generates more fluent responses compared to CCM. Nevertheless, some cases from GPT-2 merely copy contents or concepts from the given post. For example, for the third case, GPT-2 (conv) mainly discusses the concept music.

In comparison, the generated responses from our ConceptFlow are more fluent and informative than those from both CCM and GPT-2.
For example, in the third case, ConceptFlow brings associated concepts ``\textcolor[rgb]{0,0,0.7}{sound}'' and ``\textcolor[rgb]{0.5,0,0.7}{check}'' to the response generation, hopping from the grounded concepts ``\textcolor[rgb]{0.5,0.5,0}{music}'' and ``\textcolor[rgb]{0.5,0.5,0}{advice}''. Introducing these multi-hop concepts effectively improves the informativeness and diversity of generated responses.

Figure~\ref{fig:attention} presents a case study of ConceptFlow. 
The attention score $\beta^{e_i}$ and $\gamma^f$ are presented in the form of color intensity.
The ``\textcolor[rgb]{0.5,0.5,0}{championship}'' of zero-hop, ``\textcolor[rgb]{0,0,0.7}{fan}'' of one-hop and ``\textcolor[rgb]{0.5,0,0.7}{team}'' of two-hop receive more attention than others and are used to generate the response. The concept flow from ``\textcolor[rgb]{0.5,0.5,0}{fans}'' to ``\textcolor[rgb]{0,0,0.7}{fan}'' models the concept shift from user post to response. The concept flow from ``\textcolor[rgb]{0,0,0.7}{fan}'' to ``\textcolor[rgb]{0.5,0,0.7}{team}'' further describes the concept shift in response generation. In addition, some concepts, such as ``\textcolor[rgb]{0.5,0,0.7}{win}'' and ``\textcolor[rgb]{0.5,0,0.7}{pretty}'', share higher attention and may help to understand the one-hop concepts, and are filtered out when generating response by the gate $\sigma^*$ according to the relevance with conversation topic.

\subsection{Learned Attentions on Concepts}
This experiment studies the learned attention of ConceptFlow on different groups of concepts. We consider the average attention score ($\beta$ for central concepts and $\alpha$ (Appendix~\ref{appendix C}) for two-hop concepts) from all decoding steps. The probability density of the attention is plotted in Figure~\ref{fig:attention distribution}.

Figure~\ref{fig:central} shows the attention weights on central concepts. ConceptFlow effectively attends more on golden and zero-hop concepts, which include more useful information.
The attention on two-hop concepts are shown in  Figure~\ref{fig:outer}.
ConceptFlow attends slightly more on the Golden two-hop concepts than the rest two-hop ones, though the margin is smaller---the two-hop concepts are already filtered down to high-quality ones in the ConceptFlow (select) step.
\section{Conclusion and Future Work}
ConceptFlow models conversation structure explicitly as transitions in the latent concept space, in order to generate more informative and meaningful responses.
Our experiments on Reddit conversations illustrate the advantages of ConceptFlow over previous conversational systems.
Our studies confirm that ConceptFlow's advantages come from the high coverage latent concept flow, as well as its graph attention mechanism that effectively guides the flow to highly related concepts.
Our human evaluation demonstrates that ConceptFlow generates more appropriate and informative responses while using much fewer parameters.

In future, we plan to explore how to combine knowledge with pre-trained language models, e.g. GPT-2, and how to effectively and efficiently introduce more concepts in generation models.
\subsubsection*{Acknowledgments}
Houyu Zhang, Zhenghao Liu and Zhiyuan Liu is supported by the National Key Research and Development Program of China (No. 2018YFB1004503) and the National Natural Science Foundation of China (NSFC No. 61772302, 61532010). We thank Hongyan Wang, Shuo Wang, Kaitao Zhang, Si Sun, Huimin Chen, Xuancheng Huang, Zeyun Zhang, Zhenghao Liu and Houyu Zhang for human evaluations.
\newpage
\balance
\bibliography{acl2020}
\bibliographystyle{acl_natbib}
\clearpage
\nobalance

\begin{appendices}
\section{Supplementary Results}
This section provides the supplementary results of the overall performance and ablation study to further illustrate the effectiveness of ConceptFlow.

\subsection{Supplementary Results for Overall Experiments}\label{appendix A.1}

This part presents more evaluation results of the overall performance of ConceptFlow from two aspects: relevance and novelty.

Table~\ref{tab: appendix rel} shows supplementary results on Relevance between generated responses and golden responses. ConceptFlow outperforms other baselines with large margins among all evaluation metrics. Concept-PPL is the Perplexity that calculated by the code from previous work~\cite{zhou2018commonsense}. \citet{zhou2018commonsense} calculates Perplexity by considering both words and entities. It is evident that more entities will lead to a better result in terms of Concept-PPL because the vocabulary size of entities is always smaller than word vocabulary size.

More results for model novelty evaluation are shown in Table~\ref{tab: post relevance-1}. These supplementary results compare the generated response with the user post to measure the repeatability of the post and generated responses. A lower score indicates better performance because the repetitive and dull response will degenerate the model performance. ConceptFlow presents competitive performance with other baselines, which illustrate our model provides an informative response for users.

These supplementary results further confirm the effectiveness of ConceptFlow. Our model has the ability to generate the most relevant response and more informative response than other models.
\begin{table*}[t]
\centering

\begin{tabular}{l|c|c|c|c|c|c|c}
\hline  \hline
Model     & Bleu-1 & Bleu-2 & Bleu-3 & Nist-1 & Nist-2 & Nist-3 & Concept-PPL \\ \hline
Seq2Seq   & 0.1702 & 0.0579 & 0.0226 & 1.0230  & 1.0963 & 1.1056 & - \\ \hline
MemNet    & 0.1741 & 0.0604 & 0.0246 & 1.0975 & 1.1847 & 1.1960 & 46.85 \\ \hline
CopyNet   & 0.1589 & 0.0549 & 0.0226 & 0.9899 & 1.0664 & 1.0770 & 40.27 \\ \hline
CCM       & 0.1413 & 0.0484 & 0.0192 & 0.8362 & 0.9000    & 0.9082 & 39.18 \\ \hline
GPT-2 (lang) & 0.1705 & 0.0486 & 0.0162 & 1.0231 & 1.0794 & 1.0840 & - \\ \hline
GPT-2 (conv)     & 0.1765 & 0.0625 & 0.0262 & 1.0734 & 1.1623 & 1.1745 & - \\ \hline
ConceptFlow & \textbf{0.2451} & \textbf{0.1047} & \textbf{0.0493} & \textbf{1.6137} & \textbf{1.7956} & \textbf{1.8265} & 26.76 \\ \hline  \hline
\end{tabular}
\caption{More Metrics on Relevance of Generated Responses. The relevance is calculated between the generated response and the golden response. Concept-PPL is the method used for calculating Perplexity in CCM~\citep{zhou2018commonsense}, which combines the distribution of both words and concepts together. The Concept-PPL is meaningless when utilizing different numbers of concepts (more concepts included, better Perplexity shows).}\label{tab: appendix rel}

\qquad

\begin{tabular}{l|c|c|c|c|c|c|c}
\hline  \hline  & \multicolumn{7}{c}{Novelty w.r.t. Input($\downarrow$)} \\
\hline
Model     & Bleu-1 & Bleu-2 & Bleu-3 & Nist-1 & Nist-2 & Nist-3 & Rouge-1 \\ \hline
Seq2Seq   & 0.1855 & 0.0694 & 0.0292 & 1.2114 & 1.3169 & 1.3315 & \textbf{0.1678}\\ \hline
MemNet    & 0.2240  & 0.1111 & 0.0648 & 1.6740  & 1.9594 & 2.0222 & 0.2216 \\ \hline
CopyNet   & 0.2042 & 0.0991 & 0.056 & 1.5072 & 1.7482 & 1.7993 & 0.2104 \\ \hline
CCM       & \textbf{0.1667} & 0.0741 & 0.0387 & \textbf{1.1232} & \textbf{1.2782} & \textbf{1.3075} & 0.1953 \\ \hline
GPT-2 (lang) & 0.2124 & 0.0908 & 0.0481 & 1.5105 & 1.7090 & 1.7410 & 0.1817 \\ \hline
GPT-2 (conv)     & 0.2537 & 0.1498 & 0.1044 & 1.9562 & 2.4127 & 2.5277 & 0.2522 \\ \hline
ConceptFlow & 0.1850  & \textbf{0.0685} & \textbf{0.0281} & 1.3325 & 1.4600   & 1.4729 & 0.1777 \\ \hline  \hline
\end{tabular}
\caption{More Metrics on Novelty of Generated Responses. The novelty is calculated between the generated response and the user utterance, where lower means better.}\label{tab: post relevance-1}

\end{table*}
\begin{table*}[t]
\centering
\begin{tabular}{l|c|c|c|c|c|c|c|c}
\hline  \hline
Version     & Bleu-1 & Bleu-2 & Bleu-3 & Bleu-4 & Nist-1 & Nist-2 & Nist-3 & Nist-4 \\ \hline
Base  & 0.1705 & 0.0577 & 0.0223 & 0.0091 & 0.9962 & 1.0632 & 1.0714 & 1.0727 \\ \hline
Rand & 0.1722 & 0.0583 & 0.0226 & 0.0092 & 1.0046 & 1.0726 & 1.0810  & 1.0823 \\ \hline
Distract & 0.1734 & 0.0586 & 0.0230 & 0.0097 & 1.0304 & 1.0992 & 1.1081 & 1.1096 \\ \hline
Full  & \textbf{0.2265} & \textbf{0.0928} & \textbf{0.0417} & \textbf{0.0195} & \textbf{1.4550} & \textbf{1.6029} & \textbf{1.6266} & \textbf{1.6309} \\ \hline \hline
\end{tabular}

\caption{\label{tab: appendix ablation relevance}The Generation Quality of Different Outer Hop Concept Selectors. Both Bleu and Nist are used to calculate the relevance between generated responses and golden responses.}
\end{table*}

\subsection{Supplementary Results for Outer Hop Concepts} \label{appendix A.2}
The quality of generated responses from four two-hop concept selection strategies is evaluated to further demonstrate the effectiveness of ConceptFlow.

We evaluate the relevance between generated responses and golden responses,  as shown in Table~\ref{tab: appendix ablation relevance}.
\textit{Rand} outperforms \textit{Base} on most evaluation metrics, which illustrates the quality of generated response can be improved with more concepts included. \textit{Distract} outperforms \textit{Rand} on all evaluation metrics, which indicates that concepts appearing in golden responses are meaningful and important for the conversation system to generate a more on-topic and informative response. On the other hand, \textit{Full} outperforms \textit{Distract} significantly, even though not all golden concepts are included. The better performance thrives from the underlying related concepts selected by our ConceptFlow (select). This experiment further demonstrates the effectiveness of our ConceptFlow to generate a better response.

\section{Model Details of Central Flow Encoding}\label{appendix model}
This part presents the details of our graph neural network to encode central concepts.

A multi-layer Graph Neural Network (GNN)~\citep{sun2018open} is used to encode concept $e_i \in G_\text{central}$ in central concept graph:
\begin{equation}
\small
\vec{g}_{e_i} = \text{GNN} (\vec{e}_i, G_\text{central}, H),
\end{equation}
where $\vec{e}_i$ is the concept embedding of $e_i$ and $H$ is the user utterance representation set.

The $l$-th layer representation $\vec{g}_{e_i}^{\, l}$ of concept $e_i$ is calculated by a single-layer feed-forward network (FFN) over three states:
\begin{equation}
\small
 \vec{g}_{e_i}^{\, l} = \text{FFN}\left({\vec{g}_{e_i}^{\, l-1}} \circ
 {\vec{p}^{\, l-1}} \circ 
 {\sum_{r} \sum_{e_j} f_{r}^{e_j\rightarrow e_i}\left(\vec{g}_{e_j}^{\, l-1}\right)}\right),
\end{equation}
where $\circ$ is concatenate operator. $\vec{g}_{e_j}^{\, l-1}$ is the concept $e_j$'s representation of $(l-1)$-th layer. $\vec{p}^{\, l-1}$ is the user utterance representation of $(l-1)$-th layer. 

The $l$-th layer user utterance representation is updated with the zero-hop concepts $V^{0}$:
\begin{equation}
\small
\vec{p}^{\, l-1} = \text{FFN}(\sum_{e_i \in V^{0}} \vec{g}_{e_i}^{\, l-1}).
\end{equation}
$f_{r}^{e_j\rightarrow e_i}(\vec{g}_{e_j}^{\, l-1})$ aggregates the concept semantics of relation $r$ specific neighbor concept $e_j$. It uses attention $\alpha_{r}^{e_j}$ to control concept flow from $e_i$:
\begin{equation}
\small
f_{r}^{e_j\rightarrow e_i}(\vec{e}_j^{\, l-1})=\alpha_{r}^{e_j} \cdot \text{FFN}(\vec{r} \circ \vec{g}_{e_j}^{\, l-1}),
\end{equation}
where $\circ$ is concatenate operator and $\vec{r}$ is the relation embedding of $r$. The attention weight $\alpha_{r}^{e_j}$ is computed over all concept $e_i$'s neighbor concepts according to the relation weight score and the Page Rank score~\citep{sun2018open}:
\begin{equation}
\small
\alpha_{r}^{e_j} =\text{softmax}({\vec{r}} \cdot \vec{p}^{\, l-1}) \cdot \text{PageRank}(e_j^{\, l-1}),
\end{equation}
where $\text{PageRank}(e_j^{\, l-1})$ is the page rank score to control propagation of embeddings along paths starting from $e_i$~\citep{sun2018open} and $\vec{p}^{\, l-1}$ is the $(l-1)$-th layer user utterance representation.

The $0$-th layer concept representation $\vec{e}_i^{\, 0}$ for concept $e_i$ is initialized with the pre-trained concept embedding $\vec{e}_i$ and the $0$-th layer user utterance representation $\vec{p}^{\, 0}$ is initialized with the $m$-th hidden state $h_m$ from the user utterance representation set $H$. 
The GNN used in ConceptFlow establishes the central concept flow between concepts in the central concept graph using attentions.

\section{Concept Selection} \label{appendix C}
With the concept graph growing, the number of concepts is increased exponentially, which brings lots of noises. Thus, a selection strategy is needed to select high-relevance concepts from a large number of concepts.
This part presents the details of our concept selection from ConceptFlow (select).

The concept selector aims to select top K related two-hop concepts based on the sum of attention scores for each time $t$ over entire two-hop concepts:
\begin{equation} \label{equ 18}
\alpha_{n} =\sum_{t=1}^{n} \text{softmax} (\vec{s}_t \cdot \vec{e}_k),
\end{equation}
where $\vec{s}_t$ is the $t$-th time decoder output representation and $\vec{e}_k$ denotes the concept $e_k$'s embedding.

Then two-hop concepts are sorted according to the attention score $\alpha_{n}$. In our settings, top 100 concepts are reserved to construct the two-hop concept graph $V^2$.
Moreover, central concepts are all reserved because of the high correlation with the conversation topic and acceptable computation complexity.
Both central concepts and selected two-hop concepts construct the concept graph $G$.

\end{appendices}
\end{document}